\documentclass[
]{ceurart}

\usepackage{listings}
\usepackage{enumitem}

\begin{document}

\copyrightyear{2022}
\copyrightclause{Copyright for this paper by its authors.
  Use permitted under Creative Commons License Attribution 4.0
  International (CC BY 4.0).}

\conference{The 2nd World Conference on eXplainable Artificial Intelligence}

\title{Privacy Implications of Explainable AI in Data-Driven Systems}


\author[1,2]{Fatima Ezzeddine}[%
email=fatima.ezzeddine@usi.ch,
]
\address[1]{Università della Svizzera italiana, Lugano, Switzerland}
\address[2]{Scuola universitaria professionale della Svizzera italiana, Lugano, Switzerland}




\begin{abstract}
Machine learning (ML) models, demonstrably powerful, suffer from a lack of interpretability. The absence of transparency, often referred to as the black box nature of ML models, undermines trust and urges the need for efforts to enhance their explainability. Explainable AI (XAI) techniques address this challenge by providing frameworks and methods to explain the internal decision-making processes of these complex models. Techniques like Counterfactual Explanations (CF) and Feature Importance play a crucial role in achieving this goal. Furthermore, high-quality and diverse data remains the foundational element for robust and trustworthy ML applications. In many applications, the data used to train ML and XAI explainers contain sensitive information. In this context, numerous privacy-preserving techniques can be employed to safeguard sensitive information in the data, such as differential privacy. Subsequently, a conflict between XAI and privacy solutions emerges due to their opposing goals. Since XAI techniques provide reasoning for the model behavior, they reveal information relative to ML models, such as their decision boundaries, the values of features, or the gradients of deep learning models when explanations are exposed to a third entity. Attackers can initiate privacy breaching attacks using these explanations, to perform model extraction, inference, and membership attacks. This dilemma underscores the challenge of finding the right equilibrium between understanding ML decision-making and safeguarding privacy. 
\end{abstract}
\vspace{0pt}
\maketitle
\vspace{-15pt}
\section{Context and Motivation} \label{introduction}
In recent years, advancements in Artificial Intelligence (AI) have expanded beyond the primary objective of predictive capabilities. Although accurate predictions are crucial, an equally important goal has emerged: ensuring explainability. Explainability in Machine Learning (ML) models has become a critical objective for making clear and justifiable predictions, especially in high-stakes social decisions. It is essential for these models to offer clear and comprehensible reasons for their predictions and decisions \cite{jordan2015machine, lipton2018mythos}. In this context, Explainable AI (XAI) has emerged as a crucial field of investigation. XAI methodologies are specifically designed to unveil the decision-making processes of complex, opaque models, often referred to as black boxes. With the use of XAI techniques, researchers can gain valuable insights into the reasoning behind model decisions, after they have already been made \cite{holzinger2019causability, ribeiro2016should}. XAI techniques employ various methods to interpret the inner workings of complex ML models. These methods generate different types of explanations, e.g., feature importance, counterfactual explanations, etc. To generate tailored explanations, XAI requires a combination of data, interpretable models, and explanatory techniques and often incorporates user interaction. Therefore, XAI starts with the foundational element of data, which needs to be diverse and of high quality. This data is not only used to train AI models but also to create explainers. This combination of data, interpretable models, explanatory techniques, and user interaction builds the XAI.

In many applications, the data used to train AI and XAI models contain sensitive information about individuals, such as medical records, or financial transactions, which the GDPR \cite{regulation2018general} seeks to safeguard. Different approaches are proposed to safeguard sensitive information in data, such as differential privacy (DP) and federated learning (FL). These approaches affect predictive performance to some extent, resulting in a drop in performance, yet they manage to uphold an acceptable level of it. Subsequently, a conflict between ensuring transparency through XAI and ensuring privacy emerges due to their opposing goals. XAI aims to provide insights into model behavior for transparency, while privacy-preserving solutions obscure data to prevent data leakage. Moreover, the output of XAI can unintentionally expose model decision boundaries, leading to potential attacks on privacy \cite{shokri2021privacy, goethals2023privacy}. For instance, attackers can exploit XAI explanations such as CFs, which describe the minimal feature value change to alter the model decision and return instances that are close to the decision boundary. FI, which scores the contribution and impact of each feature to the model output exposes information about the gradients in Deep Neural Networks (DNNs) or about the values of the features in ML. In this context, attackers can initiate attacks from these explanations to perform model extraction, inference, and membership attacks \cite{rigaki2020survey, kim2018informational}, especially when the model is shared or deployed publicly on the cloud as ML as a Service (MLaaS). This dilemma underscores the challenge of finding the right equilibrium between explainability and safeguarding private information \cite{shokri2021privacy}.

\section{Background on Explainable Artificial Intelligence}
\subsection{Motivation and Definition}
In order to enhance transparency, XAI techniques provide the necessary tools to open up complex black boxes and shed light on how AI decisions are made \cite{gunning2017explainable, speith2022review}, promoting fairness, transparency, and accountability within real-world organizations \cite{bunn2020working}. Moreover, XAI has proven to play a pivotal role in ensuring that AI is trusted and used responsibly. By answering essential "How?" and "Why?" inquiries regarding AI systems, XAI serves as a valuable tool for tackling the increasing ethical and legal issues associated with them. XAI targets diverse entities and includes various stakeholders, such as researchers, model developers like engineers and data scientists, as well as practitioners.

\subsection{Post-hoc Explainability}
Post-hoc explainability is a technique used to gain insight into the decision-making process of a trained ML model. In this context, post-hoc means that the model's interpretability is addressed after its training, regardless of its complexity or the algorithms used. The approach primarily revolves around the act of querying the model with diverse sets of input data to observe how it reacts to different scenarios. Through these interactions, we can effectively map out the decision boundaries the model uses, shedding light on what factors influence its predictions.

Visualizations and explanations can then be applied to make these insights more accessible and human-friendly, ultimately enabling a better understanding of the model's predictions. These visual aids are essential in making the insights gained more accessible to data scientists, end users, and domain experts who are willing to understand why the model is making specific predictions. By going through this process, post-hoc explainability serves a vital role in improving model transparency and building trust in its performance.
\vspace{0pt}
Understanding an AI system with XAI relies on its training data, process, and model. Therefore, XAI can be applied throughout the entire AI development pipeline. Specifically, it can be applied in different stages of modeling, such as before, during, and after (post-modeling explainability). In this work, the primary emphasis will be on post-modeling XAI (Post-hoc), since ML models are often developed with only predictive performance in mind.

\emph{\textbf{Feature Importance}}
Feature Importance (FI) explanations involve assigning a quantitative measure in the form of a numerical score to each input feature within a given model. The primary goal of calculating FI is to discern which features have influential effects on the model's predictions and which ones have a relatively lesser impact. These importance scores help practitioners and data scientists gain insights into which factors are most critical in influencing those decisions. Features that, when modified, cause more substantial shifts in the model's output are considered more important because they have a greater influence on the final prediction. For deep learning models, many feature-based explanation functions are gradient-based techniques that analyze the gradient flow through a model. Approaches such as Layer-wise Relevance Propagation (LRP) \cite{bach2015pixel}, Integradient gradients (IG) \cite{sundararajan2017axiomatic}, and Deep Learning Important FeaTures (DeepLIFT) \cite{shrikumar2017learning} exist. %
For model-agnostic FI frameworks, perturbation methods exist such as Shapley explanations (SHAP) \cite{NIPS2017_7062}, local Interpretable Model-agnostic Explanations (LIME) \cite{ribeiro2016should}.

\emph{\textbf{Counterfactual Explanations}}
CFs leverage the concept of potential outcomes to assess causal relationships within a data-model framework. CFs empower informed decision-making and the implementation of explainable, accountable, and ultimately more ethically responsible AI \cite{wachter2017counterfactual}. It achieves this by constructing a hypothetical scenario, distinct from the observed data, and evaluating the corresponding model output under this scenario. The generation of informative and interpretable CFs necessitates the optimization of well-defined metrics \cite{verma2020counterfactual} such as diversity, validity, proximity, and user constraints. Conversely, model-specific methods tailor the cost function optimization process to leverage the inherent characteristics of the employed model. For instance, in the case of differentiable models, gradients play a critical role in guiding the optimization towards finding CFs \cite{selvaraju2017grad, wang2020scout}. Various other methods have been proposed to optimize the cost function \cite{ustun2019actionable,tolomei2017interpretable,dhurandhar2018explanations}. Conversely, model-agnostic methods achieve generalization across diverse model architectures. This is facilitated by their reliance on optimizing a cost function solely based on input/output pairs, independent of the underlying model's internal structure \cite{moore2019explaining,goodfellow2014explaining, kanamori2021ordered,nguyen2021counterfactual,ezzeddine2023sac}.
\vspace{-5pt}


\section{Related Work: Interplay between XAI and Privacy}
\subsection{Context and Problem Formulation}
\begin{figure}[t]
\includegraphics[scale=0.35]{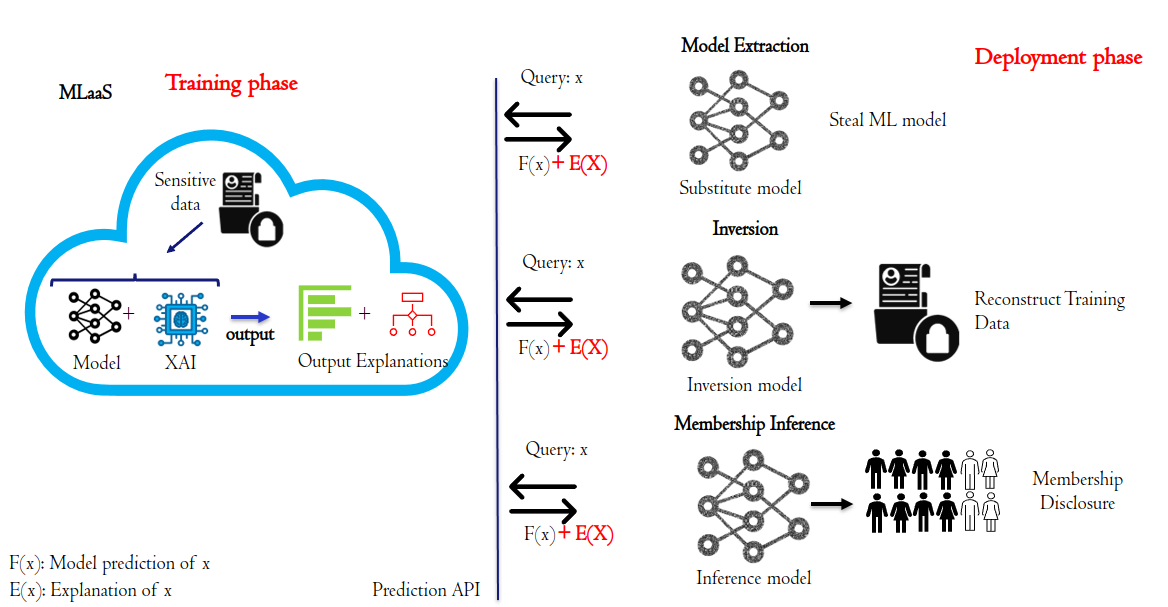}
\caption{Scenario of privacy attacks where MLaaS provides explanations alongside the prediction}
\label{fig:scenario}
\end{figure}
\textit{Data protection} and \textit{privacy} is one of the primary dimensions in ML and AI. It involves ensuring that the data used to train and test ML models does not expose sensitive information about individuals or entities. This is particularly critical when dealing with datasets that contain personally identifiable information or confidential details. Techniques like anonymization and DP have emerged as valuable tools in the data privacy field. They allow us to protect the privacy of individuals represented in the data, even as we leverage it to train models. Beyond data privacy, \textit{model privacy} is also a pressing concern. The architecture of ML models can be susceptible to privacy breaches. Models may unintentionally encode information about the training data they were exposed to, and this could pose risks when shared or deployed publicly on the cloud as MLaaS \cite{pajola2021fall}. Attacks such as model extraction, inversion, or membership inference can exploit these vulnerabilities (Details in the following sections and Fig. \ref{fig:scenario}). However, privacy is not included in the default behavior of most ML algorithms. They tend to learn not just the general trends but also the specifics of the data, potentially revealing sensitive information when the model is made public.
In an ideal scenario, we want these algorithms to focus on extracting general trends and patterns from the data while deliberately avoiding the inclusion of specific details about the data. This emphasis on distilling general patterns means that the algorithms should primarily capture the fundamental, common insights that are valuable for decision-making, aligning with privacy concerns, as they identify important details without risking individual privacy. XAI can inadvertently compromise privacy by revealing sensitive information about the model's decision boundaries. Moreover, the process of returning real data points with CFs can inadvertently expose specific instances from the training set or behaviors. Also, the process of assigning FI scores exposes the values of gradients and the feature values themselves. This conflict makes striking the right balance between model explainability and data privacy crucial to ensuring that XAI enhances our understanding of AI systems without leaking individual privacy.
\vspace{0pt}
\subsection{Attacks on Machine Learning Models}

\subsubsection{Membership inference Attacks}
A membership inference attack (MIA) is a privacy-related threat in ML where an adversary attempts to determine whether a specific data point was part of the training dataset of a deployed model \cite{shokri2017membership, patel2022model}. MIA are particularly concerning because they can compromise the privacy of individuals whose data was part of the training dataset. If an attacker can determine that a specific data point was included in the training data, it may reveal sensitive information about that individual, even if the model's output does not directly disclose such information. To perform membership attacks, \cite{shokri2017membership} proposes a shadow training process that mimics the target model with shadow models, and trains the attack model using data that is extracted using data synthesis. Also, \cite{sablayrolles2019white} discusses and proves that points with a very high loss tend to be far from the decision boundary and are more likely to be non-members. Regarding how explanation can facilitate performing MIA, \cite{shokri2021privacy} quantifies information leakage in model predictions when explanations are provided. The authors evaluate feature-based explanations, highlighting how back-propagation-based explanations reveal decision boundaries.


\subsubsection{Model Extraction Attack}
Model extraction (MEA) is a class of attacks where an adversary tries to reverse-engineer a target model by observing its behavior and querying it. MEA can potentially lead to the theft of intellectual property compromising proprietary models \cite{pal2020activethief, tramer2016stealing}. Authors in \cite{tramer2016stealing} discuss the weakness in ML services that take incomplete data with confidence levels and show successful attacks on different ML models like decision trees, SVMs, and DNNs by using equation-solving, path-finding algorithms. Regarding how explanations can facilitate MEA, FIs, and CFs have proved their ability to reveal the decision boundary of a target model \cite{miura2021megex}. Several works with FI such as \cite{oksuz2023autolycus} show how LIME infers the decision boundaries.
\cite{yan2022towards} perform the attack by minimizing task‐classification loss and task‐explanation loss. Authors in \cite{milli2019model} show how gradient-based explanations quickly reveal the model itself and highlight the power of gradients. Regarding CFs, \cite{aivodji2020model} proposed an attack that relies on the CFs of the target model to train the attack model directly. Also, \cite{wang2022dualcf} proposes a strategy to target the decision boundary shift by taking not only the CF but also the CF of the CF as pairs of training samples.

\subsubsection{Model Inversion Attack}
A model inversion attack (MINA) is a privacy-related threat in ML where an adversary attempts to reconstruct sensitive or private information about individual data points from trained model predictions. In other words, the MINA task is to predict the input data, that is, the original dataset for the target model. In \cite{zhu2022label}, authors propose label-only model MINA, while \cite{zhao2021exploiting} discusses how providing explanations harms privacy and studies this risk for image-based MINA on private image data from model explanations. The authors developed several CNN architectures that achieve significantly higher inversion performance than using only the target model prediction. To minimize the risk of MINA, \cite{NEURIPS2022_70d638f3} presents a generative noise injector for model FI explanations by perturbing model explanations, and \cite{gong2023netguard} proposes to insert engineer fake samples during the training process.
\vspace{-5pt}
\section{Research Questions and Objectives}
We pose the following research questions (RQs):
\begin{enumerate}[noitemsep]
    \item To what extent does the utilization of known privacy-preserving techniques, such as DP, effectively safeguard privacy and prevent information leakage when combined with explanations provided by XAI?
    \item Can we produce high-quality XAI explanations while safeguarding privacy to mitigate potential vulnerabilities to attacks?
    \item Which approach, privacy-preserving XAI or privacy-preserving ML, offers a more effective solution for safeguarding sensitive information in XAI systems?
\end{enumerate}


To address RQ1, we aim to evaluate the trade-off and assess the effectiveness of existing privacy-preserving techniques (e.g., DP) in mitigating information leakage when combined with XAI explanations for CFs and FI. This will involve investigating the extent to which explanations can be exploited for privacy attacks like MIA, MEA, or MENA.

To address RQ2, we aim to explore the possibility of generating high-fidelity XAI explanations while simultaneously safeguarding privacy. Potential methodologies include the exploration of multi-modeling and multi-objective reinforcement learning (RL) frameworks. These approaches aim to develop an XAI framework that concurrently optimizes two objectives: \emph{i) generating high-quality CFS}, and \emph{ii) adhering to pre-defined privacy constraints}. Furthermore, the integration of DP during the backpropagation of gradients for FI computation is another promising avenue for investigation. This approach could enable the computation of privacy-preserving FIs, mitigating the risk of information leakage from the FI explanation process itself.

To address RQ3, we will conduct a comparative analysis of privacy-preserving XAI and privacy-preserving ML techniques. This analysis will evaluate their strengths and weaknesses in safeguarding sensitive information within XAI systems. By comprehensively assessing these aspects, we aim to identify the approach that offers a more robust and enduring mechanism for privacy protection within XAI applications, covering different types of data.

\section{Results and contributions to date}
In the initial research, I explored CF generation through RL, with the specific goal of constructing an explainer that operates independently of input data. The investigation then progressed to a more in-depth examination of CFs, focusing on their potential for information leakage and their ability to reveal the decision boundaries of ML models. To reach this aim, a new methodology is proposed to carry out MEA through a concept known as knowledge distillation (KD). I also delved into the domain of explainable deep learning methods within distributed systems, such as Vertical Split Learning (VSL), aiming to evaluate the potential information disclosure resulting from FI across various entities. In addition, I analyzed the impact of DP on the explainability of anomaly detection (AD) models.More specifically:
\begin{enumerate}[noitemsep]
        \item \textbf{Explored how RL can be leveraged to generate CF explanations} without relying on the dataset as input to the explainer. The main aim is to let the CF generator learn generalizable patterns from the training data without exposing it. The explainer determines which features to modify and by how much, by maximizing a custom reward function designed to jointly optimize various metrics \cite{ezzeddine2023sac}.

        \item \textbf{Designed a new attack approach to evaluate the use of KD for an MEA} in scenarios where CFs are given to an attacker. I benefit from the property of KD and the process of transferring knowledge from a large model to a smaller one. The findings reveal that employing KD with the presence of CFs can indeed yield successful MEA. 
        
        \item \textbf{Proposed an approach to generate private CFs} I introduce the concept of DP within the GANs CF generation pipeline to generate CFs that deviate from the statistical properties of the confidential dataset, offering a layer of protection against potential privacy breaches.

        \item \textbf{Explored VSL} strategies and performed experiments to explore the risk of information leakage regarding the original features using gradient-based explanations (IG and DeepLIFT). My application of VSL focused on a use case related to Network Function Virtualization. My findings highlight how an attacker on the server side can exploit XAI techniques to achieve additional tasks, without access to the original features \cite{ezzeddine2023vertical}.

        \item \textbf{Explored DP with AD} Analyzed the trade-off between privacy achieved by DP and explainability achieved using SHAP.
\end{enumerate}

\section{Expected next steps and final contribution to knowledge}

This PhD research aims to achieve significant advancements in bridging the critical gap between XAI and data privacy. We will address the inherent conflict between providing users with clear explanations of AI models and protecting their sensitive data (privacy). We aim to develop a defense mechanism in the form of high-quality explanations while simultaneously ensuring privacy. And also bridges the gap by conducting a comparative analysis of Privacy-Preserving techniques in XAI and privacy-preserving ML.

\newpage
{\footnotesize \bibliography{sample-ceur}}
\end{document}